%% file: main.tex
\documentclass[fleqn,10pt]{wlscirep}
\usepackage[utf8]{inputenc}
\usepackage[T1]{fontenc}
\usepackage{amssymb}
\usepackage{pifont}
\usepackage{algorithm}
\usepackage{algpseudocode}
\usepackage[utf8]{inputenc}
\usepackage{amsmath}
\usepackage{multicol}
\usepackage{graphicx}
\usepackage{subcaption}
\usepackage{caption}
\usepackage{float}
\usepackage{soul}
\usepackage[markup=underlined]{changes}
\usepackage{xcolor}

\definecolor{darkgreen}{rgb}{0.0, 0.5, 0.0}

\title{Enhanced Federated Anomaly Detection Through Autoencoders Using Summary Statistics-Based Thresholding}

\author[1,*]{Sofiane Laridi}
\author[1]{Gregory Palmer}
\author[2]{Kam-Ming Mark Tam}

\affil[1]{L3S Research Center, Faculty of Electrical Engineering and Computer Science, Leibniz University Hannover, 30167, Germany}
\affil[2]{Department of Architecture, University of Hong Kong, Knowles Building, Pokfulam Road, Hong Kong SAR}
\affil[*]{laridi@l3s.de}

\begin{abstract}
In Federated Learning (FL), anomaly detection (AD) is a challenging task due to the decentralized nature of data and the presence of non-IID data distributions. This study introduces a novel federated threshold calculation method that leverages summary statistics from both normal and anomalous data to improve the accuracy and robustness of anomaly detection using autoencoders (AE) in a federated setting. Our approach aggregates local summary statistics across clients to compute a global threshold that optimally separates anomalies from normal data while ensuring privacy preservation. We conducted extensive experiments using publicly available datasets, including Credit Card Fraud Detection, Shuttle, and Covertype, under various data distribution scenarios. The results demonstrate that our method consistently outperforms existing federated and local threshold calculation techniques, particularly in handling non-IID data distributions. This study also explores the impact of different data distribution scenarios and the number of clients on the performance of federated anomaly detection. Our findings highlight the potential of using summary statistics for threshold calculation in improving the scalability and accuracy of federated anomaly detection systems.

\end{abstract}

\begin{document}
\flushbottom
\maketitle

\thispagestyle{empty}

\section*{Introduction}
\input{introduction.tex}

\section*{Literature Review}
\input{related-work.tex}

\section*{Problem Definition}
\input{problem-definition.tex}

\section*{Method}
\input{methodology.tex}

\section*{Results}
\input{experimental-results.tex}

\section*{Conclusion \& Future Work}
\input{conclusion.tex}

\section*{Data Availability}
The datasets analyzed during the current study are publicly available as follows:
\begin{itemize}
    \item \textbf{Credit Card Fraud Detection}: Provided by Worldline and the Machine Learning Group of ULB, 2014. This dataset can be accessed at \href{https://www.kaggle.com/mlg-ulb/creditcardfraud}{Kaggle}.
    \item \textbf{Shuttle Dataset}: Available from the UCI Machine Learning Repository. This dataset can be accessed at \href{https://archive.ics.uci.edu/dataset/148/statlog+shuttle}{UCI Shuttle Dataset}.
    \item \textbf{Covertype Dataset}: Available from the UCI Machine Learning Repository. This dataset can be accessed at \href{https://archive.ics.uci.edu/dataset/31/covertype}{UCI Covertype Dataset}.
\end{itemize}

\bibliographystyle{unsrt}
\bibliography{bibliography.bib}
\end{document}

%% file: introduction.tex
In Federated Learning (FL), the effectiveness of Anomaly Detection (AD) is crucial due to the decentralized nature of data across multiple clients. Federated Autoencoders (FAE) have emerged as a popular approach for detecting anomalies in such settings, as they allow each client to train local models on their data and collaboratively learn a global model without sharing raw data. However, determining an optimal threshold for AD within this federated framework remains a significant challenge.

Traditional AD methods often rely on thresholds determined from local validation data by using the trained AE to predict the validation data and thus the reconstruction errors of this validation data, then calculating a threshold that separates anomalies from normal data. These methods fail to capture the global data distribution in federated environments, leading to sub-optimal performance (McMahan et al.\cite{mcmahan2017communication}). They rely on local thresholds that may not adequately reflect the variations in data distributions across FL clients, especially when the data is non-IID (Kairouz et al.\cite{kairouz2021advances}). Additionally, many existing methods focus solely on normal data to establish the threshold, neglecting the characteristics of anomalies, compromising their effectiveness in certain scenarios.

To overcome these limitations, this study introduces an FL threshold calculation method that integrates summary statistics from both normal and anomalous data across local validation datasets of all clients, producing a federated threshold that learns the decision boundary more accurately and robustly. Research in this direction has been limited. This study demonstrates that incorporating both summary statistics and anomalies in the threshold determination process enhances the accuracy and robustness of AD in FL (Yang et al.\cite{yang2019federated}).  The research compares the performance of the proposed FL threshold calculation technique against conventional local and federated methods across various FL and data distribution scenarios, examining different numbers of clients, degrees of data distribution skewness, and anomaly rates. The findings also provide insights into the method's effectiveness across diverse FL environments and data distribution conditions.  

In summary, this paper makes the following contributions: \begin{itemize} \item Proposes a novel approach for calculating the FL threshold for AD using FL clients' local summary statistics and aggregated global statistics, while preserving privacy. \item Collects and compares several state-of-the-art threshold calculation techniques, thoroughly analyzing their performance under different FL and data distribution scenarios. \item Offers insights into whether a client should adopt the proposed FL threshold or rely on a locally calculated threshold based on the comparison of local and global summary statistics. \end{itemize}

%% file: related-work.tex
FL is widely used in AD due to its effectiveness in handling distributed data while maintaining privacy. Various techniques have been integrated into FL frameworks to detect anomalies. On the other hand, AEs have been effective in detecting malicious network activities (Li et al.\cite{li2020federated}). In IoT networks, FL helps collaboratively train models for AD, with additional security provided by blockchain integration (Ali et al.\cite{ali2021integration}). Multi-task learning within FL, discussed by Smith et al.\cite{smith2017federated}, addresses task heterogeneity across devices by leveraging diverse data sources. The FedGroup model, for example, computes learning updates from a group of devices, demonstrating its effectiveness in anomaly detection in IoT environments (Li et al.\cite{li2020federated}).

% Autoencoders and their role in FAD
However, federated AD faces several challenges. A major challenge is handling non-IID data across clients, which can negatively impact the model's performance (Bonawitz et al.\cite{bonawitz2019towards}). Traditional methods often assume IID data, but newer approaches, such as graph-based methods by Liu et al.\cite{liu2008isolation}, consider data relationships for better detection. Zhao et al.\cite{zhao2018federated} have focused on optimizing FL systems to deal with non-IID data and improve communication efficiency. Generative Adversarial Networks are used to identify anomalies through reconstruction errors or discriminator scores (Schlegl et al.\cite{schlegl2019f}). Other methods, such as softmax scores from neural network classifiers by Hendrycks \& Gimpel\cite{hendrycks2016baseline} or outlier exposure by Liang et al.\cite{liang2017enhancing}, help improve AD. Combining generative and discriminative models by Nalisnick et al.\cite{nalisnick2019detecting} has also been shown to enhance accuracy. Other advanced neural architectures, like attention-based models by Vaswani et al.\cite{VaswaniSPUJGKP17} and LSTM networks for temporal data by Zhu et al.\cite{zhu2019long}, are used.

% local thresholding techniques 
In addition, \textcolor{black}{AEs} are effective tools for AD, leveraging their ability to reconstruct input data and highlight deviations as anomalies (Xu et al.\cite{xu2018unsupervised}). By learning compact representations, AEs excel in identifying unusual patterns across various domains, including network security and IoT systems (Li et al.\cite{li2020federated}). More complex versions of AEs, such as Variational Autoencoders, enhance this process by modeling probabilistic distributions, improving the detection of subtle anomalies (An \& Cho\cite{an2015variational}). Different adaptations of AEs address specific data types: convolutional autoencoders for spatial anomalies (Baur et al.\cite{baur2021autoencoders}) and hybrid models combining AEs with LSTM networks for temporal data (Malhotra et al.\cite{malhotra2015long}). Attention mechanisms integrated into AEs by Zong et al.\cite{zong2018deep} further refine AD by focusing on the most relevant data features. Additionally, combining AEs with Generative Adversarial Networks (GANs) has shown robust performance in \textcolor{black}{AD} through enhanced reconstruction quality (Zenati et al.\cite{zenati2018efficient}). Despite all the mentioned AE variations, fully connected AEs are preferred for handling large-scale data due to their simplicity, interpretability, and efficiency (Hinton \& Salakhutdinov\cite{hinton2006reducing}). 

Given their reliance on reconstruction error, effective AD with AEs requires appropriate threshold calculation techniques, which are crucial aspects of AD. In FL, this task is particularly challenging due to the distributed nature of data across multiple clients. Therefore, different threshold calculation methods for AEs have been explored. The majority of methods we have found calculate the threshold locally by focusing on client-specific data without federated aggregation. For instance, the Local Kernel Quantile Estimator (KQE) (Huong et al.\cite{KQE}) sets thresholds based on reconstruction error quantiles using a kernel estimator. The Local IQR Range method determines thresholds using the interquartile range of reconstruction errors. Similarly, the Local Percentile method (Percentile) uses a specific percentile of the error distribution to establish thresholds (Novoa-Paradela et al.\cite{fastAE}). Other local methods include the Local Largest MSE (Largest-MSE) (Sáez-de-Cámara et al.\cite{saez2023clustered}), which bases thresholds on the highest observed MSE. The Local Peak Over Threshold (POT) method identifies anomalies by focusing on errors that exceed a high quantile (Kea et al.\cite{kea2023enhancing}). Schlegl et al.\cite{schlegl2017unsupervised} calculate a threshold by generating multiple thresholds between the minimum and maximum reconstruction errors of the validation data (Local-MinMax), then selecting the threshold with the highest F1 score.

Federated threshold calculation techniques have been explored as well. One common approach is where Wang et al.\cite{AvgLocalThreshold} set a federated threshold by averaging the mean-squared error plus the standard deviation across clients (Fed-MSE-StD). While straightforward, this method can be less effective since it does not consider anomalies directly in the threshold calculation process. Sánchez et al.\cite{sanchez2022studying} calculate the mean and standard deviation of the local thresholds and filter out thresholds with a z-score greater than 1.5 (Fed-Filtered). The global threshold is then set as the maximum of the remaining filtered thresholds. One of the limitations of these federated techniques is the neglect of actual anomalies in the federated threshold calculation. 
An alternative federated approach by Pourahmadi et al.\cite{pourahmadi2022spotting} involves generating candidate thresholds between the global minimum and maximum reconstruction errors (Fed-MinMax), with clients selecting the optimal threshold based on F1 scores. However, this interval between the global minimum and maximum can be vague, especially in non-IID data distributions, and manually setting the number of thresholds might not capture the optimal threshold. Additionally, this method does not suggest a fair aggregation of the threshold candidates' F1 scores, which gives equal influence to clients with large validation data and clients with minimal validation data. This can lead to a sub-optimal global threshold. 

The limitations of these methods underscore the need for a more automated and equitable threshold calculation method that can account for the diverse data distributions across clients and provide a balanced approach in FL. We assume that summary statistics, such as mean, variance, skewness, and kurtosis, can offer valuable insights into the distribution of the clients' validation data, helping to develop thresholds that are more representative of the overall data landscape across FL clients.
One of the significant advantages of using summary statistics in FL is their privacy-preserving nature. Unlike raw data, which contains detailed information about individual data points, summary statistics aggregate this information into a form that reveals general trends without exposing specific data values (McMahan et al.\cite{mcmahan2017communication}).

We have listed in Table \ref{tab:threshold_methods} the different state-of-the-art methods mentioned, along with their characteristics and limitations, for better clarification.

\begin{table}[h]
\centering
\small % Make the font size smaller to fit the page
\begin{tabular}{|l|c|c|c|c|}
\hline
\textbf{Thresholding Method} & \textbf{Federated} & \textbf{Anomalies} & \textbf{Statistics Used} & \textbf{Local Data Distribution Consideration} \\ \hline
\textbf{Our Method}          & \ding{51}          & \ding{51}                   & Mean, Variance, etc. & \ding{51} \\ \hline
Fed-MinMax                   & \ding{51}          & \ding{51}                   & Min/Max              & \ding{55} \\ \hline
Fed-MSE-StD                  & \ding{51}          & \ding{55}                   & Mean, StD            & \ding{55} \\ \hline
Fed-Filtered                 & \ding{51}          & \ding{55}                   & Mean, StD            & \ding{55} \\ \hline
Local-MinMax                 & \ding{55}          & \ding{51}                   & MSE, Percentile      & \ding{55} \\ \hline
KOE                          & \ding{55}          & \ding{55}                   & ---                  & \ding{55} \\ \hline
IQR                          & \ding{55}          & \ding{55}                   & IQR                  & \ding{55} \\ \hline
Percentile                   & \ding{55}          & \ding{55}                   & Percentile           & \ding{55} \\ \hline
Largest-MSE                  & \ding{55}          & \ding{55}                   & MSE                  & \ding{55} \\ \hline
POT                          & \ding{55}          & \ding{55}                   & High Quantile        & \ding{55} \\ \hline
Local-MSE-Std                & \ding{55}          & \ding{55}                   & Mean, StD            & \ding{55} \\ \hline
\end{tabular}
\caption{SOTA Threshold Calculation Approaches}
\label{tab:threshold_methods}
\end{table}

%% file: problem-definition.tex
Federated AD using \textcolor{black}{FAE} aims to collaboratively identify anomalies in distributed data without sharing raw data between clients. The key challenge is to determine an optimal global threshold $\theta_{\text{global}}$ for AD, leveraging each client's local validation data while preserving privacy.

\subsection*{Notations}

\begin{multicols}{2}
\begin{itemize}
    \setlength{\itemsep}{0pt} % Adjust the space between items
    \setlength{\parskip}{0pt} % Adjust the space between paragraphs
    \item $D_{\text{train},i}$: Training dataset for client $i$
    \item $M_t$: Global FAE model at round $t$
    \item $M_{t,i}$: Local model for client $i$ at round $t$
    \item $D_{\text{val},i}$: Validation dataset for client $i$ containing both normal and anomalous samples
    \item $\hat{D}_{\text{val},i}$: Reconstructed validation dataset for client $i$
    \item $E_i$: Array of reconstruction errors for client $i$, where $E_i = \{e_{i,1}, e_{i,2}, \ldots, e_{i,n}\}$
    \item $\mu_i$: Mean of the reconstruction errors for client $i$
    \item $\sigma_i^2$: Variance of the reconstruction errors for client $i$
    \item $S_i$: Skewness of the reconstruction errors for client $i$
    \item $K_i$: Kurtosis of the reconstruction errors for client $i$
    \item $N_i$: Number of samples in the validation dataset for client $i$
    \item $\mu_{\text{global}}$: Global mean of the reconstruction errors
    \item $\sigma^2_{\text{global}}$: Global variance of the reconstruction errors
    \item $S_{\text{global}}$: Global skewness of the reconstruction errors
    \item $K_{\text{global}}$: Global kurtosis of the reconstruction errors
    \item $T$: Array of $n$ thresholds, $T = \{\mu_1, \mu_2, \ldots, \mu_n\}$ determined based on the overlap region
    \item $F_{i}$: Array of F1 scores for client $i$ corresponding to thresholds $T$
    \item $F_{\text{avg}}$: Array of average F1 scores for each threshold across all clients
    \item $\theta_{\text{global}}$: Global threshold with the highest average F1 score
\end{itemize}
\end{multicols}

\subsection*{Given}

\begin{enumerate}
    \setlength{\itemsep}{0pt} % Adjust the space between items
    \setlength{\parskip}{0pt} % Adjust the space between paragraphs
    \item A FAE model $M$ trained on distributed training data $D_{\text{train}}$.
    \item Each client's local validation data $D_{\text{val},i}$ with corresponding reconstruction errors $E_i$.
\end{enumerate}

\subsection*{Objective}

To compute a global threshold $\theta_{\text{global}}$ that optimally separates anomalies from normal data across all clients while addressing the following challenges:

\begin{itemize}
    \setlength{\itemsep}{0pt} % Adjust the space between items
    \setlength{\parskip}{0pt} % Adjust the space between paragraphs
    \item \textbf{Data Privacy}: Each client's validation data $D_{\text{val},i}$ is private and cannot be shared with other clients or the server.
    \item \textbf{Data Distribution}: Each client may have different data distributions, leading to variations in reconstruction errors $E_i$.
\end{itemize}

%% file: methodology.tex
Our methodology for federated anomaly detection using autoencoders is divided into two main steps: Federated Autoencoder Training and Federated Threshold Calculation. The detailed steps of our methodology are outlined below:

\subsection*{Federated Auto-Encoder Training}

The first step involves training a \textcolor{black}{FAE} model using the Federated Averaging (FedAvg) algorithm to aggregate the clients' model weights, as illustrated in Algorithm 1. \textcolor{black}{Model synchronization between clients and the server occurs at the beginning of each round, where the server broadcasts the global model to all clients, and at the end of each round, clients send their updated weights back to the server for aggregation, following the FedAvg paradigm.} We selected a fully connected AE due to its proven effectiveness in \textcolor{black}{AD}, particularly for high-dimensional datasets such as Shuttle, Covertype, and Credit Card Fraud Detection (\textcolor{black}{Sakurada \& Yairi}\cite{sakurada2014anomaly}).

\begin{algorithm}
\caption{Federated \textcolor{black}{Auto-Encoder} Training with FedAvg Aggregation}
\begin{algorithmic}[1]
\State \textbf{Input:} $D_{\text{train},i}$: Training dataset for client $i$, $N$: Number of clients, $E$: Local epochs, $\eta$: Learning rate, $R$: Rounds
\State \textbf{Output:} $M$: Trained global autoencoder model
\State Initialize global model $M_0$
\For{each round $t = 1, \ldots, R$}
    \State Server sends global model $M_{t-1}$ to all clients
    \For{each client $i = 1, \ldots, N$ \textbf{in parallel}}
        \State Initialize local model $M_{t,i}$ with $M_{t-1}$ weights
        \State Train $M_{t,i}$ on $D_{\text{train},i}$ for $E$ epochs using learning rate $\eta$
        \State Send local updates $\Delta M_{t,i}$ to server
    \EndFor
    \State Server aggregates updates to form global model $M_t$
    \State Update global model: $M_{t} \leftarrow M_{t-1} + M_t$
\EndFor
\State \textbf{Return:} Trained global model $M_T$
\label{algo-1}
\end{algorithmic}
\end{algorithm}

\begin{figure}[h]
\centering
\includegraphics[width=0.6\textwidth]{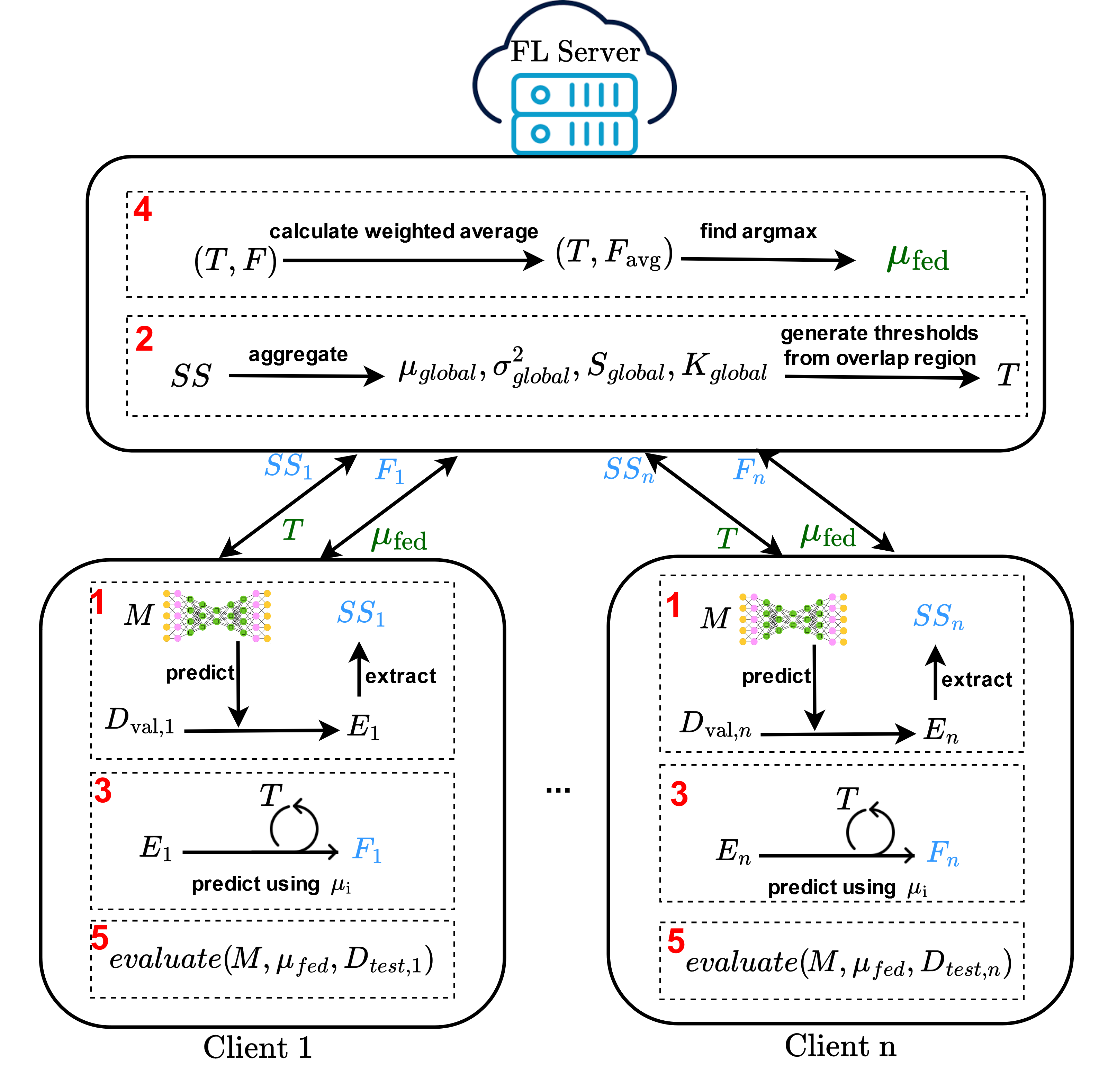}
\caption{Federated Threshold Calculation using Summary Statistics}
\label{fig:fed-thresholding}
\end{figure}

\subsection*{Summary Statistics-Based Threshold Selection}

Figure \ref{fig:fed-thresholding} illustrates the workflow of our federated \textcolor{black}{AD} approach using \textcolor{black}{FAE}, emphasizing how thresholds are calculated using aggregated summary statistics. The process is divided into several key steps:

\subsubsection*{Prediction and Extraction of Summary Statistics}

Each client employs its local Autoencoder (AE) model \( M \) to compute reconstruction errors \( E_i \) on its local validation dataset \( D_{\text{val},i} \). Based on these errors, the client calculates summary statistics \( SS_i \), including the mean (\(\mu_i\)), variance (\(\sigma_i^2\)), skewness (\(S_i\)), and kurtosis (\(K_i\)).

\subsubsection*{Weighted Aggregation of Summary Statistics and Threshold Selection}

After aggregating the summary statistics from all clients, the server identifies the overlap region where the distributions of normal and anomalous reconstruction errors intersect. This overlap region aids in isolating reconstruction error outliers by focusing on thresholds generated within this specific range.

The initial overlap region is determined using the global means (\( \mu_{\text{normal}} \), \( \mu_{\text{anomaly}} \)) and standard deviations (\( \sigma_{\text{normal}} \), \( \sigma_{\text{anomaly}} \)) of the normal and anomalous distributions, calculated via the upper and lower bounds. These bounds are subsequently fine-tuned by adjusting for skewness and kurtosis to ensure that the region accurately reflects the data's distributional shape. Skewness compensates for asymmetry in the data, while kurtosis adjusts the bounds to account for tail behavior.

\subsubsection*{Threshold Candidates Generation}

Within the refined overlap region, the server generates a set of candidate thresholds. These candidates are typically spaced evenly across the overlap range, with the number of candidates determined by a predefined parameter (e.g., 1000 candidates). The generated thresholds are then distributed to all clients for evaluation.

\subsubsection*{F1 Score Calculation and Aggregation}

Each client evaluates these candidate thresholds using its local validation data, calculating the F1 score for each threshold. The server aggregates these F1 scores across all clients to select the global threshold \( \theta_{\text{global}} \), which maximizes the overall F1 score.

The detailed steps of this method are illustrated in Algorithm 2.

\begin{minipage}[t]{0.48\textwidth}
\begin{algorithm}[H]
\caption{Federated Threshold with Weighted Aggregation}
\begin{algorithmic}[1]
\setlength{\itemsep}{0pt} % Reduce space between lines
\setlength{\parskip}{0pt} % Reduce space between paragraphs
\State \textbf{Input:} $D_{\text{val},i}$: Validation dataset for client $i$
\State \textbf{Output:} $\mu_{\text{global}}$: Global threshold
\For{each client $i$}
    \State Use the trained model $M$ to reconstruct $D_{\text{val},i}$.
    \State Compute reconstruction errors $E_i$ for each sample in $D_{\text{val},i}$.
    \State Calculate summary statistics for $E_i$: mean $\mu_i$, variance $\sigma_i^2$, skewness $S_i$, kurtosis $K_i$, and count $N_i$.
\EndFor
\For{each client $i$}
    \State Send summary statistics $\mu_i$, $\sigma_i^2$, $S_i$, $K_i$, and $N_i$ to the server.
\EndFor
\setlength{\itemsep}{0pt} % Reduce space between lines
\setlength{\parskip}{0pt} % Reduce space between paragraphs
\State Server computes the global summary statistics of both normal and anomaly using weighted aggregation:
\State \hspace{0.5cm} $\mu_{\text{global}} = \frac{\sum_{i=1}^{k} N_i \mu_i}{\sum_{i=1}^{k} N_i}$
\State \hspace{0.5cm} $\sigma^2_{\text{global}} = \frac{\sum_{i=1}^{k} N_i \left(\sigma_i^2 + (\mu_i - \mu_{\text{global}})^2 \right)}{\sum_{i=1}^{k} N_i}$
\State \hspace{0.5cm} $S_{\text{global}} = \frac{\sum_{i=1}^{k} N_i S_i \cdot \sqrt{N_i} \cdot \left(\frac{\sigma_{\text{global}}}{\sigma_i}\right)^3}{\sum_{i=1}^{k} N_i}$
\State \hspace{0.5cm} $K_{\text{global}} = \frac{\sum_{i=1}^{k} N_i K_i \cdot N_i \cdot \left(\frac{\sigma_{\text{global}}}{\sigma_i}\right)^4}{\sum_{i=1}^{k} N_i}$
\end{algorithmic}
\end{algorithm}
\end{minipage}%
\hfill
\begin{minipage}[t]{0.48\textwidth}
\begin{algorithm}[H]
\label{algo:2}
\begin{algorithmic}[1]
\setcounter{ALG@line}{15}
\State Server determines the overlap region:
\State \hspace{0.5cm} $\text{Lower Bound} = \max(\mu_{\text{normal}} - 3\sigma_{\text{normal}}, \mu_{\text{anomaly}} - 3\sigma_{\text{anomaly}})$
\State \hspace{0.5cm} $\text{Upper Bound} = \min(\mu_{\text{normal}} + 3\sigma_{\text{normal}}, \mu_{\text{anomaly}} + 3\sigma_{\text{anomaly}})$
\State Server generates an array of $n$ thresholds $T = \{\mu_1, \mu_2, \ldots, \mu_n\}$ within the overlap region.
\State Server sends the threshold array $T$ to each client.
\For{each client $i$}
    \For{each threshold $\mu_j$ in $T$}
        \State Calculate F1 scores $F_i$ for each threshold $\mu_j$.
    \EndFor
\EndFor
\For{each client $i$}
    \State Send F1 score array $F_i$ to the server.
\EndFor
\State Server calculates average F1 scores $F_{\text{avg}}$ for each threshold.
\State Server identifies the threshold $\mu_{\text{global}}$ with the highest average F1 score in $F_{\text{avg}}$.
\State Server sends the global threshold $\mu_{\text{global}}$ to all clients.
\end{algorithmic}
\end{algorithm}
\end{minipage}

%% file: experimental-results.tex
In this study, we evaluate our federated thresholding method using three publicly available datasets: Credit Card Fraud Detection (284,807 samples, 492 anomalies, 29 dimensions), Shuttle (49,097 samples, 3,511 anomalies, 9 dimensions), and Covertype (581,012 samples, 2,747 anomalies, 10 dimensions). The datasets were sourced from UCL and Kaggle. The following steps outline our data preparation and distribution for simulating a federated learning environment.

\begin{itemize}

\item \textbf{Normalization and Scaling:} All datasets were normalized and scaled to ensure consistency across features. This preprocessing step is crucial for the performance of the AE, enabling it to effectively reconstruct normal data for AD across different clients.

\item \textbf{Train-Validation-Test Splitting:} Each dataset was split into training, validation, and test sets. The training data consisted solely of normal samples, as the AE is trained to model normal data behavior. Both validation and test sets included a mixture of normal and anomalous data, which were used to assess the thresholding methods. For scalability experiments, we varied the number of clients from 2 to 50, distributing the data evenly across clients to study the effect of client count on model performance (Bonawitz et al.\cite{bonawitz2019towards}).

\item \textbf{Federated Learning Splitting on Clients:} 
    To explore various federated learning scenarios:
    \begin{itemize}
        \item \textbf{Evenly Distributed Data:} The training, validation, and test data were uniformly distributed across all clients. This uniform distribution served as a baseline to assess our method under ideal and balanced conditions, which is a standard assumption in many federated learning experiments (Li et al.\cite{li2020federated}).
        
        \item \textbf{Non-IID Data:} For the Shuttle and Covertype datasets, which include multiple classes, we designated one class as anomalous while treating the remaining classes as normal data. The normal data was distributed across clients, with each client receiving data from different classes. The anomalous data was divided among clients using the k-means clustering algorithm to ensure a diverse distribution of anomalies. Since these datasets include seven classes, we assigned six clients for normal data and distributed the anomalous data among them. To maintain consistency, the Credit Card Fraud Detection dataset, which is binary, was also split among six clients by applying the k-means clustering algorithm (Ahmed et al.\cite{ahmed2020k}) to both normal and anomalous data, thus simulating a highly non-IID scenario similar to that of the other datasets (Zhao et al.\cite{zhao2018federated}).
    \end{itemize}

\end{itemize}

\begin{table}[ht]
\scriptsize % Further decrease the overall font size
\centering
\begin{minipage}{0.48\textwidth}
\centering
\begin{tabular}{lccc}
\toprule
\textbf{Method} & \textbf{Shuttle} & \textbf{Credit Card} & \textbf{Cover} \\
\midrule
\textbf{Our Method} & \textbf{0.9873} & \textbf{0.8985} & \textbf{0.8440} \\
Fed Threshold & 0.9861 & 0.8972 & 0.8395 \\
Fed Mean MSE + StD & 0.9844 & 0.8959 & 0.4003 \\
Fed Filtered Threshold & 0.9865 & 0.8881 & 0.3637 \\
Local Iterative & 0.9836 & 0.8878 & 0.8228 \\
Local Inter Quantile Range & 0.9773 & 0.8862 & 0.7318 \\
Local Percentile & 0.9657 & 0.8865 & 0.5273 \\
Local Kernel Quantile Estimator & 0.9775 & 0.8768 & 0.4512 \\
Local Max MSE & 0.9849 & 0.8594 & 0.3584 \\
Local Mean MSE + Std & 0.9619 & 0.8817 & 0.4828 \\
Local Peak Over Threshold & 0.9802 & 0.8125 & 0.4551 \\
\bottomrule
\end{tabular}
\caption{Average F1 Scores Across Different Methods Using Evenly Distributed Data}
\label{tab:f1_scores_even}
\end{minipage}\hfill
\begin{minipage}{0.48\textwidth}
\centering
\begin{tabular}{lccc}
\toprule
\textbf{Method} & \textbf{Shuttle} & \textbf{Credit Card} & \textbf{Cover} \\
\midrule
\textbf{Our Method} & \textbf{0.9251} & \textbf{0.8725} & 0.8351 \\
Fed Threshold & 0.9033 & 0.8705 & 0.8383 \\
Fed Mean MSE + StD & 0.3841 & 0.8556 & 0.7121 \\
Fed Filtered Threshold & 0.4231 & 0.8460 & 0.7542 \\
Local Iterative & 0.9120 & 0.8712 & \textbf{0.8558} \\
Local Inter Quantile Range & 0.4628 & 0.8416 & 0.7605 \\
Local Percentile & 0.4742 & 0.8419 & 0.7890 \\
Local Kernel Quantile Estimator & 0.4781 & 0.8347 & 0.8027 \\
Local Max MSE & 0.3546 & 0.8143 & 0.7803 \\
Local Mean MSE + Std & 0.4614 & 0.8363 & 0.8088 \\
Local Peak Over Threshold & 0.3615 & 0.7938 & 0.7822 \\
\bottomrule
\end{tabular}
\caption{Average F1 Scores Across Different Methods Using Non-IID Data}
\label{tab:f1_scores_noniid}
\end{minipage}
\end{table}

\begin{figure}[!htb]
    \centering
    \includegraphics[width=\textwidth]{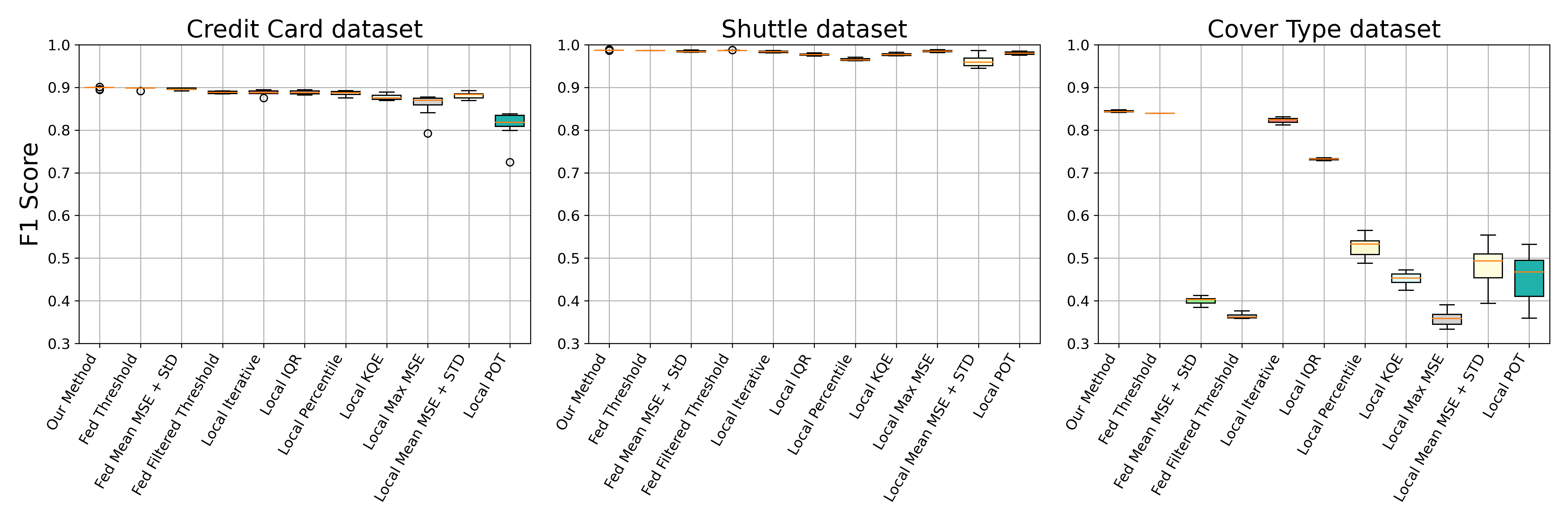}
    \caption{Comparison of F1 Scores Across Different Threshold Calculation Methods Using Randomly Distributed Data}
    \label{fig:avg_f1_scores_box_plots}
\end{figure}

\subsection*{Data Distribution}

\subsubsection*{Even Distribution}

The results demonstrate that our method for generating thresholds is effective across various data distribution scenarios. In the evenly distributed data setting (Table \ref{tab:f1_scores_even}), our method consistently achieves the highest F1 scores across all datasets, indicating superior performance in AD compared to other methods. Additionally, our method remains highly reliable, maintaining strong performance even as the number of clients increases. The \textbf{Fed Mean MSE + StD} method also performs well but is typically outperformed by our approach. In contrast, local methods such as \textbf{Local Inter Quantile Range} and \textbf{Local Percentile} exhibit greater variability in their F1 scores, rendering them less consistent and reliable in distributed data scenarios. Overall, these results confirm that federated approaches, particularly our proposed method, are highly effective in managing evenly distributed data across multiple clients.

\subsubsection*{Non-IID Distribution}

In the Non-IID data scenario (Table \ref{tab:f1_scores_noniid}), where clients have heterogeneous data distributions, our method consistently outperforms other approaches, achieving the highest F1 scores across the Shuttle (0.9251), Credit Card (0.8725), and Covertype (0.8351) datasets. Notably, the \textbf{Local-MinMax} method remains highly competitive, particularly in the Credit Card dataset where it attains an F1 score of 0.8712, nearly matching the performance of our method and surpassing other federated approaches. Similarly, in the Covertype dataset, the \textbf{Local-MinMax} method achieves an F1 score of 0.8558, outperforming all other methods, including federated ones.

These findings suggest that while the federated approach effectively aggregates information across clients to establish a robust global threshold, certain local methods, particularly \textbf{Local-MinMax} Thresholding, can more effectively address client-specific data variations. This is especially evident in datasets like Credit Card and Covertype, where data distributions vary significantly across clients.

Given these insights, it is evident that there are scenarios where a client may benefit more from utilizing a local threshold rather than a federated one. This observation underscores the importance of investigating and understanding the conditions under which a client should prefer the federated threshold over its own local threshold. Future analyses will focus on the summary and aggregated statistics utilized in our federated threshold calculation to determine the optimal thresholding strategy for different client scenarios.

\subsubsection*{Random Distribution}

To evaluate the robustness of our method under varying data distribution conditions, we employed diverse random distributions with varying numbers of samples per client. This approach allowed us to simulate a wide range of scenarios that may occur in federated learning environments. Figure \ref{fig:avg_f1_scores_box_plots} presents the boxplots generated from these experiments, illustrating the F1 scores obtained across different random setups.

The width and spread of the boxplots provide insights into the consistency and robustness of each threshold calculation method. Our proposed method is represented by consistently narrower boxplots, indicating lower variance in F1 scores across different random distribution scenarios. This narrow spread suggests that our method is both robust and reliable, maintaining high performance regardless of variability in data distribution among clients.

In contrast, the wider boxplots associated with some of the other methods indicate greater variability in performance. This variability suggests that these methods are more sensitive to changes in data distribution, resulting in less consistent outcomes. Additionally, the presence of outliers in these boxplots further highlights the instability of these methods under certain random distribution setups.

Overall, the relatively narrow boxplots of our method demonstrate its superior robustness, as it maintains high and stable F1 scores across a diverse set of random distribution scenarios. This underscores the adaptability and effectiveness of our federated thresholding approach, even under challenging and unpredictable data conditions.

\subsubsection*{Global Anomaly and Normal Reconstruction Data Overlap Analysis}
\begin{figure}[!htb]
    \centering
    \includegraphics[width=\textwidth]{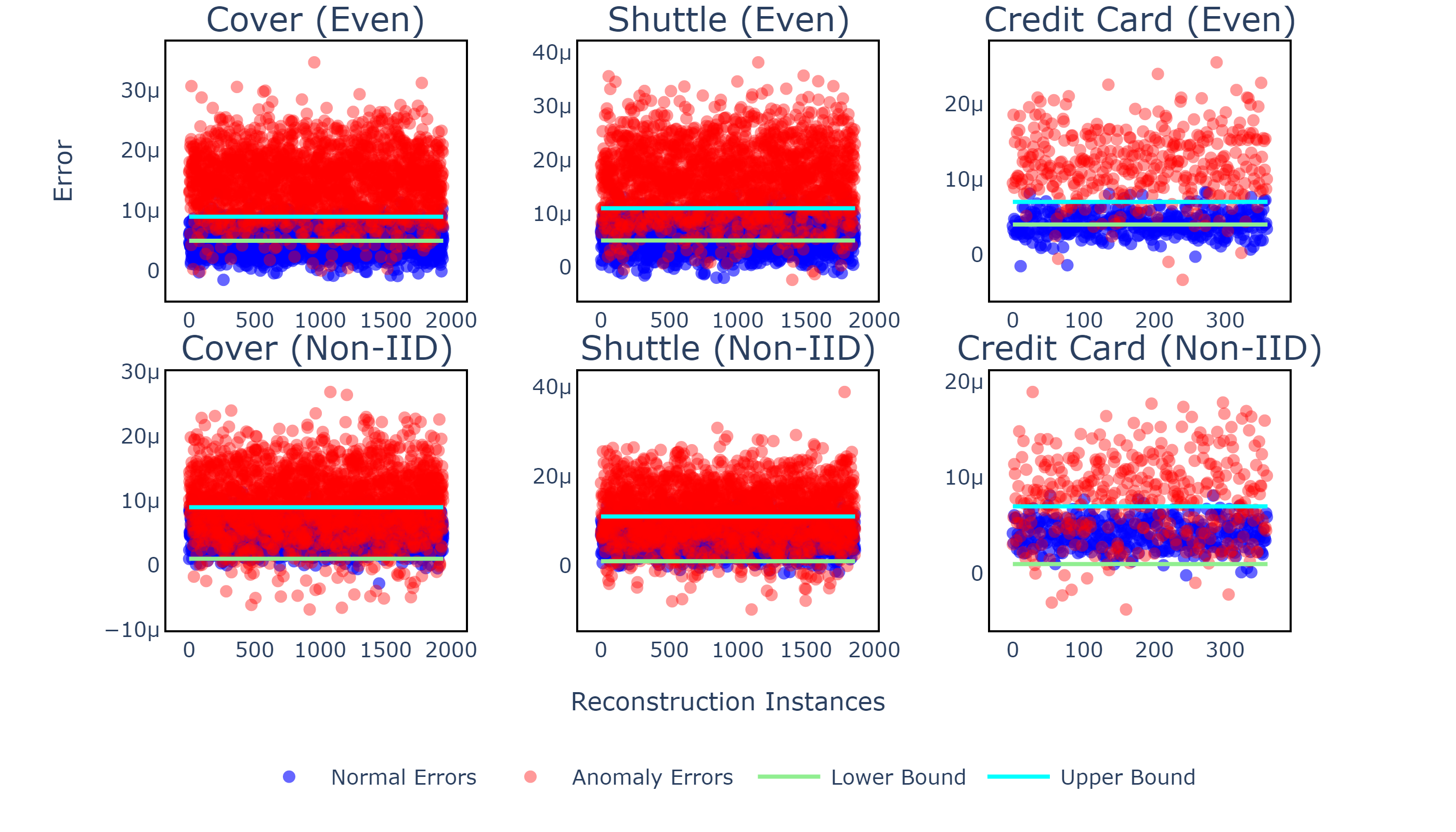}
    \caption{Visualization of Error Distribution with Overlap Regions Defined by Upper and Lower Bounds Across Different Datasets and Data Distributions}
    \label{fig:overlap}
\end{figure}

In addition to performance evaluation, the global reconstruction errors, as depicted in Figure \ref{fig:overlap}, provide valuable insights into the functioning of our method under different data distribution scenarios. The visualizations reveal that in evenly distributed data scenarios (top row), the overlap region between normal and anomalous reconstruction errors is smaller compared to Non-IID scenarios (bottom row). This smaller overlap suggests that in evenly distributed data, fewer threshold candidates are needed to accurately distinguish between normal and anomalous data points.

Conversely, the larger overlap observed in Non-IID scenarios indicates that a higher number of threshold candidates may be necessary to achieve similar accuracy. This difference highlights the necessity of adapting the threshold generation process based on the data distribution. Specifically, more candidates are likely required in Non-IID settings to manage the increased overlap between normal and anomalous errors effectively.

These findings emphasize the importance of considering data distribution characteristics when designing thresholding methods in federated learning. By accounting for the extent of overlap between normal and anomalous reconstruction errors, our method can dynamically adjust the threshold selection process to maintain high detection accuracy across diverse and complex data environments.

\subsection*{Scalability}
\subsubsection*{Number of Clients}
\begin{figure}[!htb]
    \centering
    \includegraphics[width=\textwidth]{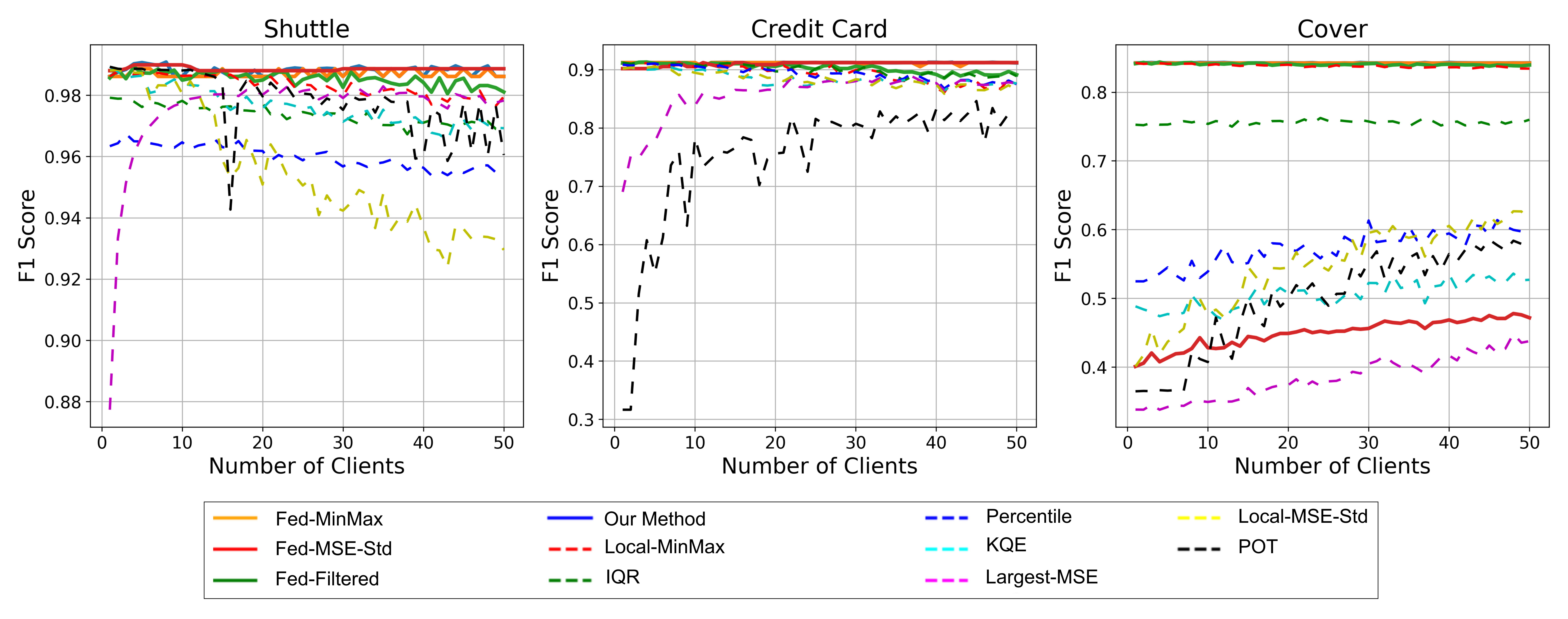}
    \caption{Impact of Number of Clients on F1 Scores Across Different Threshold Calculation Methods and Datasets}
    \label{fig:scalability}
\end{figure}
The results from the Shuttle, Covertype, and Credit Card Fraud Detection datasets, as shown in Figure \ref{fig:scalability}, demonstrate the effectiveness of various thresholding techniques. Our \textbf{Fed Threshold} method consistently achieves the highest F1 scores across all datasets and varying numbers of clients, underscoring its robustness and reliability. Specifically, for the Shuttle dataset, the \textbf{Fed Threshold} method maintains a high F1 score close to \textcolor{black}{0.99} regardless of the number of clients. The \textbf{Fed Mean MSE + StD} method also performs comparably well, suggesting that these federated approaches effectively aggregate client data to sustain performance.

In contrast, other methods such as \textbf{Local Inter Quantile Range}, \textbf{Local Percentile}, \textbf{Local Kernel Quantile Estimator}, and \textbf{Local Peak Over Threshold} generally exhibit lower F1 scores. These methods show some improvement \textcolor{black}{in the Credit Card Fraud Detection and Shuttle datasets} as the number of clients increases but fail to match the consistency and high performance of the federated approaches. \textcolor{black}{Conversely, in the Shuttle dataset, local methods—especially \textbf{Local-MSE + StD}—show a slight decrease in F1 scores as the number of clients increases. This decline occurs because, as the data is partitioned among more clients and given the extreme imbalance of the Shuttle dataset, each client receives fewer and less representative data points to calculate reliable thresholds.}

For the Covertype dataset, our \textbf{Fed Threshold} and \textbf{Fed Mean MSE + StD} methods maintain high F1 scores near 0.9, irrespective of the number of clients. Other methods, including \textbf{Local Kernel Quantile Estimator} and \textbf{Local Peak Over Threshold}, exhibit significant fluctuations and lower F1 scores. The performance of these alternative methods does not consistently improve with an increasing number of clients, highlighting the stability and effectiveness of the federated approaches.

The Credit Card Fraud Detection dataset results further reinforce these observations. The \textbf{Fed Threshold} and \textbf{Fed Mean MSE + StD} methods consistently demonstrate high F1 scores around 0.9, with minimal performance degradation as the number of clients increases. In contrast, methods such as \textbf{Local Max MSE} and \textbf{Local Peak Over Threshold} display considerable variability and generally lower F1 scores, emphasizing their instability and less effective performance in a federated setting.

Overall, the \textbf{Fed Threshold} method consistently outperforms other thresholding techniques across all datasets and client numbers. Its ability to maintain high performance and scalability makes it highly suitable for real-world scenarios where anomaly detection must be aggregated from multiple clients. The \textbf{Fed Mean MSE + StD} method also exhibits strong performance, albeit slightly less consistent than our \textbf{Fed Threshold} method.

\textcolor{black}{
When evaluating the performance of our \textbf{Fed Threshold} approach with varying numbers of clients, it remains highly effective, exhibiting minimal degradation in F1 scores as the number of clients increases. This demonstrates the robustness of the \textbf{Fed Threshold} method in maintaining accuracy, even as the client count grows, making it well-suited for real-world federated learning scenarios involving numerous clients.
}

\subsubsection*{Execution Time}

\begin{figure}[!htb]
    \centering
    \includegraphics[width=\textwidth]{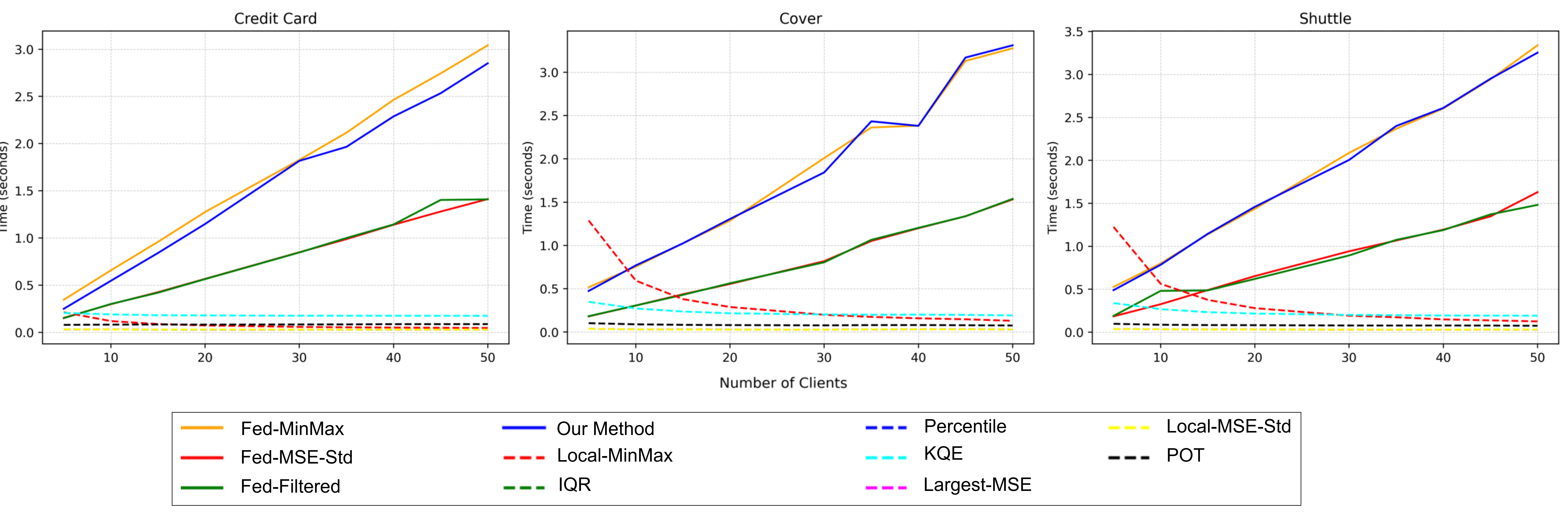}
    \caption{Execution time of different threshold calculation methods}
    \label{fig:time}
\end{figure}
The execution time results across the Credit Card Fraud Detection, Covertype, and Shuttle datasets, as shown in Figure \ref{fig:time}, reveal important trade-offs between accuracy and computational cost. The \textbf{Fed Threshold} method (solid blue line) exhibits the highest execution time, increasing significantly as the number of clients grows, reaching approximately 3 seconds for 50 clients. This increase reflects the computational complexity associated with aggregating and processing data across clients to calculate a global threshold. Although more resource-intensive, this method provides superior accuracy.

The \textbf{Fed Mean MSE + StD} method (solid red line) also shows increased execution time as the number of clients rises, though it remains slightly lower than the \textbf{Fed Threshold} method. This approach balances relatively high accuracy with more moderate computational demands. Similarly, the \textbf{Fed Filtered Threshold} method (solid green line) follows a comparable trend, indicating similar efficiency but with slightly lower accuracy.

In contrast, local methods such as \textbf{Local Iterative}, \textbf{Local IQR}, and \textbf{Local Percentile} (various dashed lines) maintain significantly lower and more stable execution times as the number of clients increases. These methods are computationally efficient because they operate on local data without the need for extensive aggregation, but they do so at the expense of lower accuracy.

Overall, federated methods such as \textbf{Fed Threshold} and \textbf{Fed Mean MSE + StD} are more computationally demanding but offer higher accuracy, making them suitable for scenarios where precision is critical. Conversely, local methods provide faster execution times, which may be preferable in resource-constrained environments or situations requiring quick decisions.

\textcolor{black}{
While the federated methods demonstrate robustness in maintaining performance as the number of clients increases, we also observed a significant increase in execution time, particularly in the Credit Card Fraud Detection dataset, where execution time grew by 460\% from 0.5 seconds to 2.8 seconds as the number of clients increased from 10 to 50. This highlights a trade-off between computational cost and accuracy, indicating that while these methods are scalable in terms of performance, their scalability in terms of computational efficiency may present challenges in resource-constrained environments.
}

\subsection*{Robustness against Noise}
\begin{figure}[!htb]
    \centering
    \includegraphics[width=\textwidth]{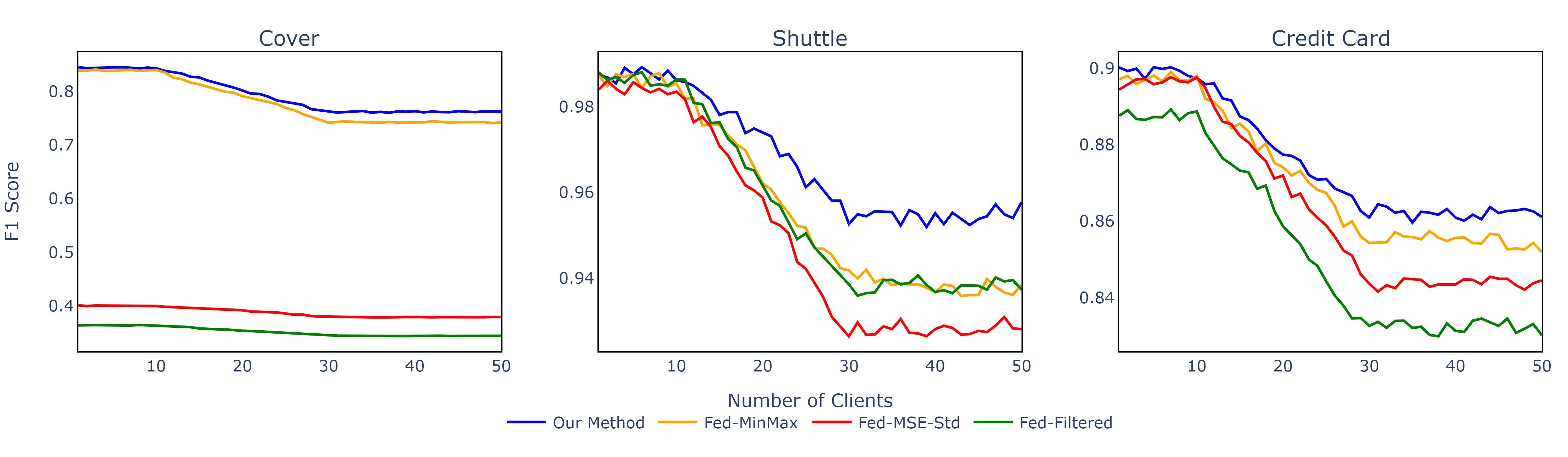}
    \caption{Impact of Corrupt Data on F1 Scores Across Different Threshold Calculation Methods with Up to 30 Corrupt Clients}
    \label{fig:noise}
\end{figure}
The robustness of the thresholding methods against noise is illustrated in Figure \ref{fig:noise}, where the x-axis represents the increasing number of clients with corrupted data, up to 30 clients.

For the Covertype dataset, \textbf{Our Method} consistently maintains a high F1 score, experiencing only a gradual decline as the number of corrupted clients increases. This demonstrates strong resilience to noise. In contrast, the \textbf{Fed-MinMax} and \textbf{Fed-Filtered} methods exhibit significantly lower F1 scores from the outset, indicating their reduced robustness to noisy data.

In the Shuttle dataset, \textbf{Our Method} again demonstrates superior robustness, maintaining a relatively high F1 score that decreases more slowly compared to other methods. Notably, the \textbf{Fed-MSE-Std} method shows a sharp decline in performance as the number of corrupted clients increases, highlighting its vulnerability to noise.

For the Credit Card Fraud Detection dataset, a similar trend is observed. \textbf{Our Method} maintains the highest F1 scores across all levels of noise, while the performance of the \textbf{Fed-Filtered} and \textbf{Fed-MSE-Std} methods decreases more rapidly. This further underscores the superior robustness of \textbf{Our Method}.

Overall, these results emphasize the robustness of \textbf{Our Method} against noise, making it a reliable choice in scenarios where data corruption is a significant concern. Other methods, while effective in certain contexts, tend to exhibit more substantial performance degradation as the number of corrupted clients increases.

\subsection*{Follow-Up Analysis: Predicting the Benefit of Federated vs. Local Threshold}

Based on observations from our previous experiments, we identified that certain clients may benefit more from using their locally calculated thresholds rather than a federated threshold. This insight prompted us to investigate whether valuable information could be extracted from the summary statistics—both local and federated—collected during those experiments. The objective of this study is to explore whether these summary statistics can be leveraged to predict when a client would benefit more from a federated threshold versus a local one.

We conducted experiments across various data distributions, varying numbers of clients, and differing levels of non-IIDness to simulate a wide range of real-world scenarios. This comprehensive approach provided a robust dataset for analysis. In each use case, we employed the Federated Autoencoder model and collected validation data to extract summary statistics. These statistics, along with the corresponding F1 scores for both local and federated thresholds, formed the foundation of our analysis.

The dataset’s features include locally calculated summary statistics from both normal and anomalous validation data, as well as aggregated summary statistics. Specifically, the local statistics comprise measures such as mean, variance, skewness, kurtosis, and count for both normal and anomalous data. The aggregated statistics are derived by combining these local statistics across all clients, producing features such as aggregated mean, variance, skewness, kurtosis, and proportional counts for both normal and anomalous data. In total, the dataset comprises 20 features.

For labeling, we focused on the difference in F1 scores between the best-performing local threshold and the federated threshold. Based on our earlier results, the \textbf{Local Iterative Threshold} was identified as the most effective local method, while our \textbf{Federated Threshold} emerged as the best federated method. Consequently, the label for each client’s data point represents the difference in F1 scores achieved by the federated threshold and the \textbf{Local Iterative Threshold} on the client's local test data.

To evaluate the benefits of federated versus local thresholds, we employed two modeling approaches: binary classification and regression. The binary classification model aimed to predict whether a client would benefit from using the federated threshold, while the regression model predicted the actual difference in F1 scores between the federated and local thresholds.

For the binary classification task, we employed a Support Vector Machine (SVM) model \cite{svm} and achieved an accuracy of 89\% when the data was randomly split into training, validation, and test sets across all three datasets. However, when we tested a more realistic scenario—training and validating on data from the Shuttle and Covertype datasets while testing on the Credit Card Fraud Detection dataset—the SVM's performance dropped to an accuracy of 68\%. This suggests that while the model performs well under random splits, its ability to generalize across different datasets is limited.

For the regression task, we employed a Random Forest model \cite{forest}. In the random split scenario, the Random Forest regressor achieved a moderate R² score of 0.67, indicating some predictive capability but also challenges in accurately predicting the differences in F1 scores. When tested under the more realistic split, the model's performance further decreased, attaining an R² score of only 0.59. This highlights the difficulty in generalizing the regression model to new, unseen datasets.

\begin{figure}[!htb]
    \centering
    \includegraphics[width=0.7\textwidth]{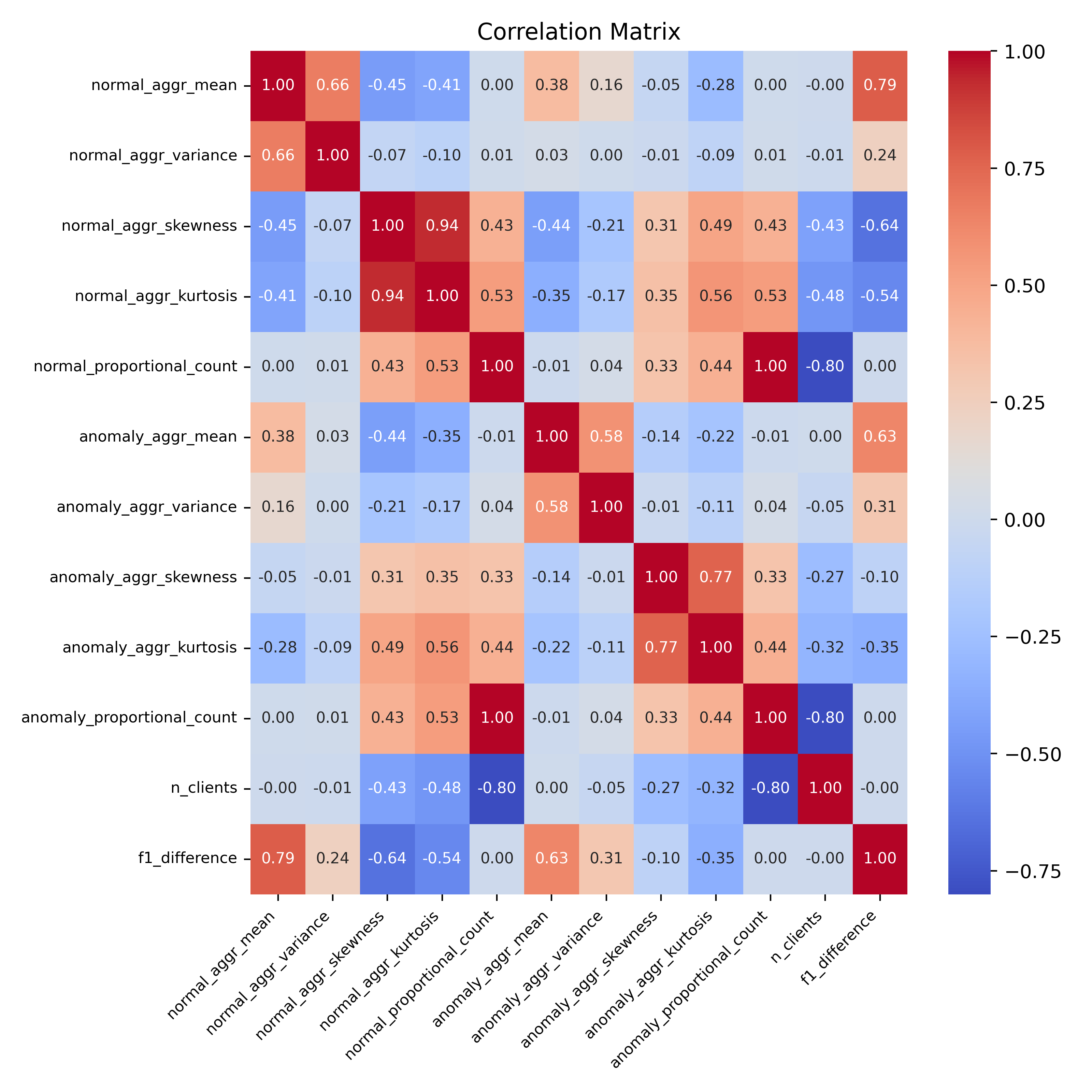}
    \caption{Correlation Matrix of Summary Statistics and F1 Score Differences in Predicting the Benefit of Federated Thresholds}
    \label{correlation}
\end{figure}

The correlation matrix presented in Figure \ref{correlation} offers further insights into the relationships between various summary statistics and the F1 score difference (\texttt{f1\_difference}). Analysis of the matrix reveals that certain features, such as \texttt{normal\_aggr\_mean} and \texttt{anomaly\_aggr\_mean}, exhibit a relatively strong correlation with \texttt{f1\_difference}. This suggests that these features are effective predictors of whether the federated threshold will outperform the local threshold. Conversely, other features, including \texttt{normal\_proportional\_count} and \texttt{anomaly\_proportional\_count}, display little to no correlation with \texttt{f1\_difference}. This indicates that these features may be less useful in predicting the advantage of employing a federated threshold.

These observations highlight opportunities for enhancing our feature selection process. By refining our feature engineering strategies and potentially incorporating more relevant features, we can improve the predictive power of our models. Additionally, the weak correlation observed for some features underscores the necessity to explore alternative statistical measures or advanced modeling techniques that can more effectively capture the complexities inherent in the data.

Our experiments emphasize the critical role of appropriate feature selection in enhancing the generalization capabilities of models within federated learning environments. The insights derived from the correlation analysis will inform future endeavors aimed at developing more adaptive and robust thresholding mechanisms, thereby advancing the effectiveness of federated anomaly detection.

%% file: conclusion.tex
In this paper, we introduced an innovative federated thresholding method for anomaly detection (AD) using autoencoders (AEs), leveraging summary statistics to improve robustness and accuracy across multiple clients. Our approach consistently outperformed traditional local and federated thresholding techniques in both IID and non-IID data scenarios. Its scalability was evident, with minimal performance degradation as the number of clients increased, making it highly suitable for real-world federated learning applications. While federated methods generally provided superior results, certain local methods remained effective in cases with highly varied data distributions.

Future research could focus on further enhancing the thresholding process by incorporating advanced statistical measures, such as entropy or mutual information, to capture more complex patterns and improve detection accuracy. Adaptive thresholding mechanisms that respond in real-time to changes in client data distributions could also increase the flexibility and responsiveness of anomaly detection systems. Additionally, integrating differential privacy techniques may further enhance data protection and ensure compliance with privacy regulations. Adaptive federated learning models that dynamically adjust to varying data distributions could further improve the robustness and scalability of federated anomaly detection systems.